\documentclass[letterpaper, 10 pt, conference]{ieeeconf}  
\IEEEoverridecommandlockouts                              

\usepackage[backend=biber,
            url=false,
            isbn=false,
            doi=false,
            backref=true,
            style=ieee,
            citestyle=numeric-comp,
            sorting=nyt,
            block=none]{biblatex}

\usepackage[keeplastbox]{flushend}
\usepackage{multirow}
\usepackage{color}
\usepackage{times}
\usepackage{biblatex}
\usepackage{balance}
\usepackage{graphicx}
\usepackage{amsmath}
\usepackage{amssymb}
\usepackage{gensymb}
\usepackage{verbatim}
\usepackage{caption}
\usepackage{hyperref} 

\newcommand{\ignore}[1]{}

\title{\LARGE \bf Fast and Reliable Autonomous Surgical Debridement with Cable-Driven Robots Using a Two-Phase Calibration Procedure}

\author{Daniel Seita$^{1}$, Sanjay Krishnan$^1$, Roy Fox$^1$, Stephen McKinley$^2$, John Canny$^{1}$, Ken Goldberg$^{1,3}$
\thanks{All authors are affiliated with AUTOLAB at the University of California, Berkeley, USA. {\tt \small http://autolab.berkeley.edu/}. $^{1}$EECS, $^2$ME, $^3$IEOR. }
}

\begin{document}
\maketitle

\begin{abstract}

Automating precision subtasks such as debridement (removing dead or diseased tissue fragments) with Robotic Surgical Assistants (RSAs) such as the da Vinci Research Kit (dVRK) is challenging due to inherent non-linearities in cable-driven systems. We propose and evaluate a novel two-phase coarse-to-fine calibration method. In Phase I (coarse), we place a red calibration marker on the end effector and let it randomly move through a set of open-loop trajectories to obtain a large sample set of camera pixels and internal robot end-effector configurations. This coarse data is then used to train a Deep Neural Network (DNN) to learn the coarse transformation bias. In Phase II (fine), the bias from Phase I is applied to move the end-effector toward a small set of specific target points on a printed sheet. For each target, a human operator manually adjusts the end-effector position by direct contact (not through teleoperation) and the residual compensation bias is recorded.  This fine data is then used to train a Random Forest (RF) to learn the fine transformation bias.  Subsequent experiments suggest that without calibration, position errors average 4.55mm.  Phase I can reduce average error to 2.14mm and the combination of Phase I and Phase II can reduces average error to 1.08mm.  We apply these results to debridement of raisins and pumpkin seeds as fragment phantoms. Using an endoscopic stereo camera with standard edge detection, experiments with 120 trials achieved average success rates of 94.5\%, exceeding prior results with much larger fragments (89.4\%) and achieving a speedup of 2.1x, decreasing time per fragment from 15.8 seconds to 7.3 seconds. Source code, data, and videos are available at \url{https://sites.google.com/view/calib-icra/}.
\end{abstract}

\section{Introduction}\label{sec:intro}

Surgical robots~\cite{surgical_robotics_2016}, such as Intuitive Surgical's da Vinci~\cite{dVRK,dvrk2014}, are widely used in laparoscopic procedures worldwide~\cite{Moustris2011,Anvari2014,Beasley2012}. Existing surgical robotics platforms rely on pure teleoperation through a master-slave interface where the surgeon fully controls the motions of the robot. To reduce tedium and fatigue in long or repetitive procedures, recent work has highlighted several opportunities for autonomous execution of surgical subtasks~\cite{robot-autonomy2017} including debridement~\cite{Kehoe2014}, suturing~\cite{Schulman2013,preda2015,sen2016automating,tsc2016}, and palpation for tumor detection~\cite{garg2016gpas}.

The da Vinci and other Robotic Surgical Assistants (RSAs) such as Applied Dexterity's Raven II~\cite{raven2013} are cable-driven and designed to be compliant as not to damage anatomy. Cable-driven robots are susceptible to issues such as cable-stretch, hysteresis, and other problems described in~\cite{pastor2013,mahler2014case,miyasaka2015,kalman_filter_2016}. 

The da Vinci Research Kit (dVRK) is a research platform built from the components of the da Vinci~\cite{dVRK,dvrk2014}. The robotic arm consists of a 3 DoF manipulator to which a modular 4 DoF instrument is attached. The tool is commanded to some position and orientation defined in the global remote center coordinate frame. However, each modular instrument has slightly different kinematics and cable tension. This can lead to errors in positional control of the tool in the global coordinate frame.

\begin{figure}[t]
\centering
\includegraphics[width=0.48\textwidth]{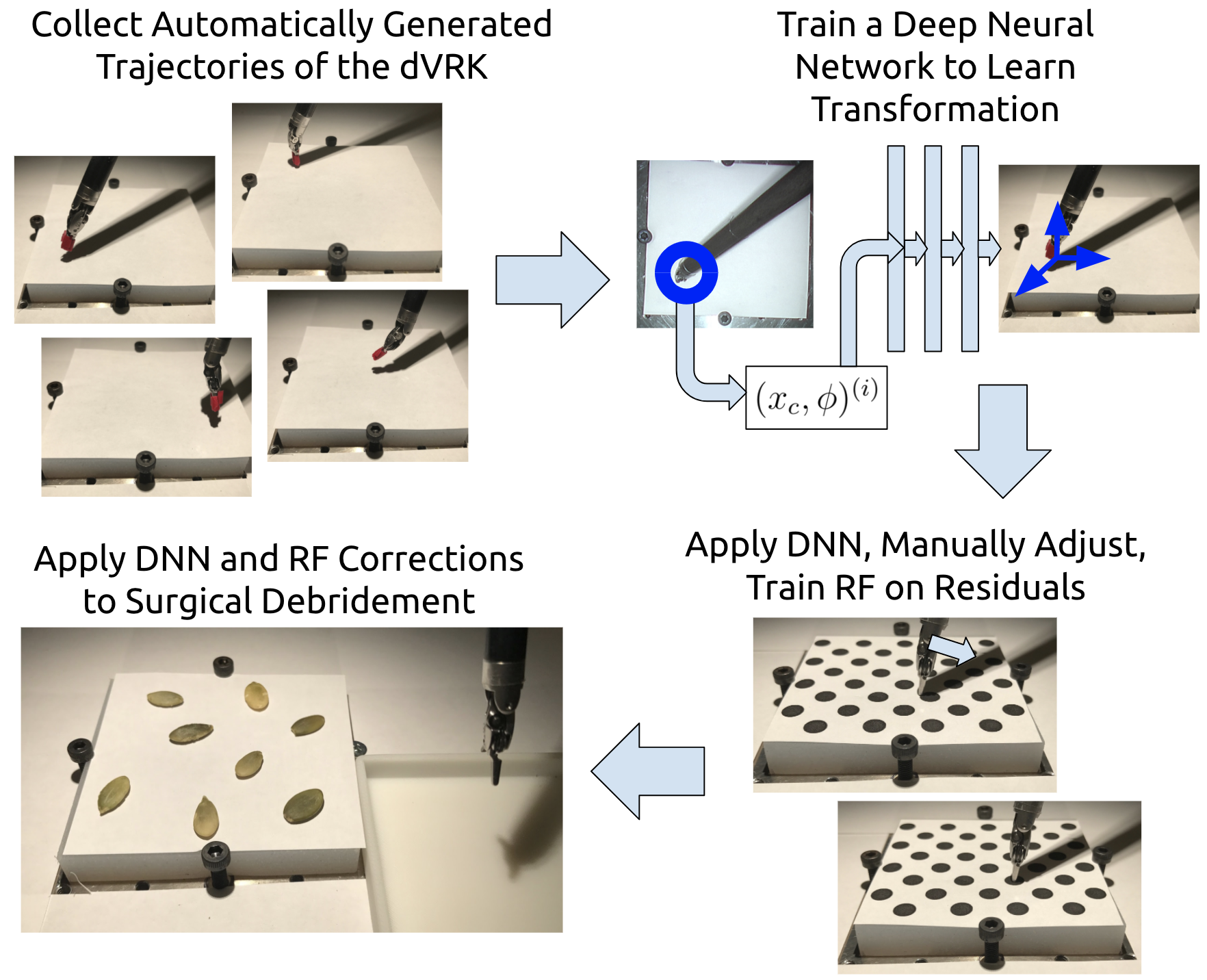}
\caption{\small
Calibration pipeline. Phase I: the dVRK, wrapped with red tape, randomly explores its workspace. The robot positions and orientations are recorded by a camera and its internal joint sensors. After data cleaning, we train a Deep Neural Network (DNN) to map from camera position (not images) and robot orientation (extracted from the images) to base position. Phase II: we apply the DNN on a calibration grid and a human directly corrects the resulting end-effector positions. The residuals are stored and used as training data for Random Forests (RFs). We apply the combined DNN and RF pipeline to debridement.
}
\label{fig:money_shot}
\end{figure}

Experiments indicate that this imprecision is systematic and repeatable, suggesting it maybe possible to anticipate and correct for bias in the commanded position of the robot. An offline calibration procedure can learn the direction and magnitude of this bias. The robot can use its internal system model as a starting point, and a spatial residual factor can be learned from visually tracking the tool in an experimental environment. 

In this paper, we show that after applying a two-phase calibration procedure shown in Figure~\ref{fig:money_shot}, the dVRK can achieve positional control in the workspace within 1mm of error while operating at the maximum speed of approximately 6cm/sec. Furthermore, the dVRK can efficiently extract fragments (sizes shown in Table~\ref{tab:dimensions}) in a surgical debridement task with 94.5\% reliability. In Phase I, the dVRK collects training data by automatically exploring trajectories in the workspace with random targets, using red tape on the end-effector to track its location. The training data input consists of tool position relative to camera frame and (discretized) $SO(3)$ rotations relative to base frame. The output consists of tool position relative to base frame. As detailed in Section~\ref{sec:problem_statement}, using rotations as input is useful in tasks such as debridement where we can determine the orientation of fragments in advance using OpenCV ellipse fitting on the camera image of the workspace. A Deep Neural Network (DNN) is trained on this data. In Phase II, we correct the DNN's errors in the predicted versus actual base frame coordinates. The dVRK's end-effector moves to target locations in a printed calibration grid, and a human directly corrects the positions. We train a Random Forest (RF) to predict these corrections.

Our approach is based on four prior papers on automated surgical debridement and/or calibration~\cite{Kehoe2014,mahler2014case,murali2015learning,pastor2013}, and extends them with three contributions: 
\begin{enumerate}
\item An alternative and cheaper calibration method compared to~\cite{pastor2013,mahler2014case,Kehoe2014}, using automatic data collection with DNNs and manual movements with RFs. We evaluate three calibration algorithms (rigid body, DNN, DNN+RF) and test various DNN settings; see Sections~\ref{sec:calibration_experiments} and~\ref{sec:debridement_experiments}.

\item Experimental results with more challenging fragments: pumpkin seeds and raisins. Sizes are shown in Table~\ref{tab:dimensions}; the fragments from~\cite{Kehoe2014,mahler2014case} (about 10mm each in width, length, and thickness) had volume about 3.3x larger and were made of foam rubber to be forgiving of gripper orientation error. Similarly, fragments from~\cite{murali2015learning} were glued onto a highly elastic surface to tolerate errors. In contrast, pumpkin seeds easily slide out of the gripper.

\item Faster execution: 2.1x speedups over prior work~\cite{mahler2014case}. 
\end{enumerate}



\section{Related Work}\label{sec:related_work}

We use the da Vinci Research Kit (dVRK)~\cite{dVRK,dvrk2014} as our RSA, which is a research platform based on Intuitive Surgical's da Vinci surgical system~\cite{dvrk_firstgen} and which has been frequently used in surgical robotics research~\cite{murali2015learning,vandenBerg2010,sen2016automating,krishnan2017ddco,thananjeyan2017}. Calibration is critical for the dVRK and other robots to reliably perform fine-grained manipulation tasks. The majority of work in calibration of kinematic parameters, reviewed in~\cite{hollerbach2008}, has focused on modeling linearities in kinematic chains. Examples include work on autonomously calibrating both a stereo vision system and robot manipulators~\cite{calib1991} and jointly calibrating multiple cameras~\cite{le2009}. Since the dVRK has inherent non-linearities, we must incorporate non-linear models in our calibration procedure.

To account for actuator imprecision, one could use visual servoing~\cite{kragic_servoing_2002} as part of an automation pipeline. For instance,~\cite{levine2017} trained a large convolutional neural network to predict the probability that specific motions would lead to successful grasps. To servo, they sampled a set of candidate commands and evaluated the likelihood of grasp success based on their trained network. In~\cite{krishnan2017ddco}, the authors applied visual servoing of the dVRK for needle insertion and picking tasks, but required large amounts of task-specific expert demonstrations to train a visual servoing policy. Learning a \emph{positional compensation} rather than an end-to-end policy may be far more sample efficient as the magnitude of the motion to learn is often smaller.

Surgical debridement~\cite{Attinger2000_CPMS,Granick2006_WRR,nichols2015} is the process of removing dead or damaged tissue to allow the remaining part to heal. Automated surgical debridement was first demonstrated in~\cite{Kehoe2014}. Using the Raven II, the authors debrided at a rate of 91.8 and 60.3 seconds per fragment for the autonomous single-arm and two-arm cases, respectively, and had success rates of 89.4\% for the single-arm. Mahler et al.~\cite{mahler2014case} used calibration to avoid replanning steps and sped up debridement by 3.8x with the two-arm setup to get a rate of 15.8 seconds per fragment with comparable success rates, but they did not report benchmarks on the slower one-arm scenario. Murali et al.~\cite{murali2015learning} used more realistic fragments but had them glued onto highly elastic tissue to tolerate several millimeters of calibration error, and they did not achieve notable speed-ups over~\cite{mahler2014case} nor did they discuss their calibration procedure.

We use a different debridement model by using small fragments with inclusion sizes that are close to the maximum width of the gripper (10mm). As an additional challenge, one of our fragment types (pumpkin seeds) is smooth and easily slides on our workstation surface. Moreover, inspired by~\cite{Kehoe2014}'s timing comparisons involving human execution, we focus on optimizing speed jointly with accuracy. These factors require high precision in control.

\section{Problem Statement}\label{sec:problem_statement}

\begin{figure}[t]
\centering
\includegraphics[width=0.35\textwidth]{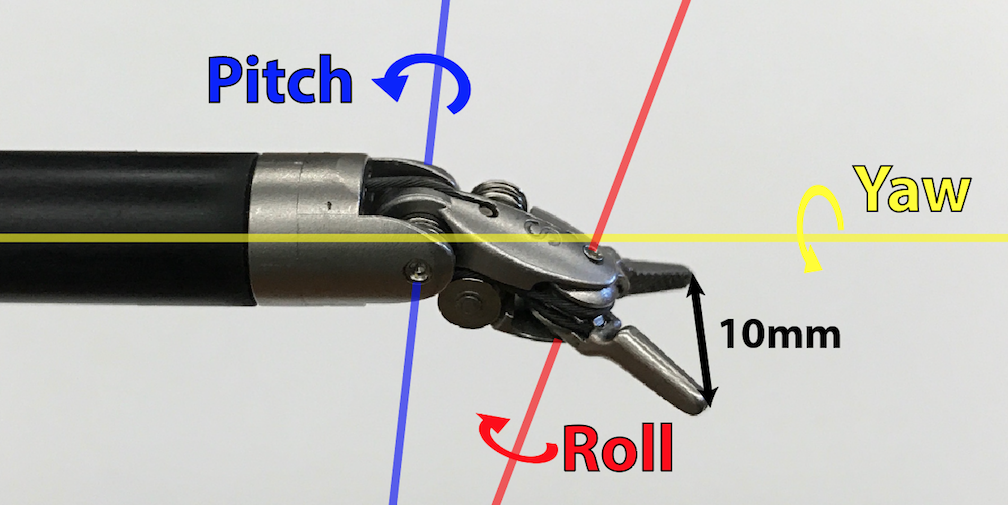}
\caption{\small
The dVRK end-effector (``large needle driver'') and its three axes of rotation. We denote these as $\phi_y$ (yaw), $\phi_p$ (pitch), and $\phi_r$ (roll). The gripper tips are opened to their maximum width (10mm).
}
\label{fig:yaw_pitch_roll}
\end{figure}


\subsection{Definitions and Notation}

Throughout the paper, we use the following notation:

\begin{itemize}
\item The \emph{camera position} $x_c = (c_x, c_y, c_z) \in \mathbb{R}^3$ is the end-effector (i.e., tool) position with respect to the \emph{camera} frame. These are determined from left and right camera pixels and their disparity.

\item The \emph{base position} $x_b = (b_x, b_y, b_z) \in \mathbb{R}^3$ is the end-effector position with respect to the \emph{base} frame.

\item The \emph{robot orientation} $\phi = (\phi_y, \phi_p, \phi_r) \in \mathbb{R}^3$ is the yaw, pitch, and roll of the end-effector in the \emph{base} frame. See Figure~\ref{fig:yaw_pitch_roll} for how the axes are oriented for our dVRK instrument (the orientation is non-standard).
\end{itemize}

\subsection{Assumptions}

\begin{figure}[t]
\centering
\includegraphics[width=0.45\textwidth]{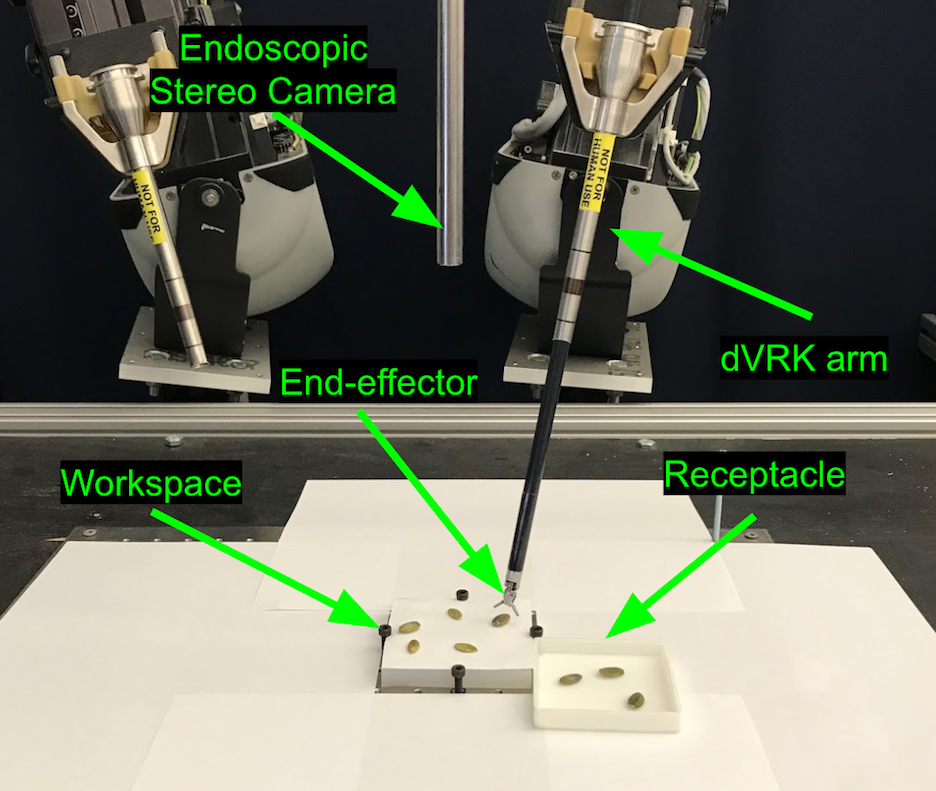}
\caption{\small
Overall view of our setup. We use the dVRK arm labeled above for our experiments. The endoscopic stereo camera contains two separate lenses and is located 7.5 inches above the $7.5\mathrm{cm}\times 7.5\mathrm{cm}$ workspace.
}
\label{fig:dvrk_setup}
\end{figure}

%

%
\begin{table}[t]
\centering
\small
\begin{tabular}{l l l l}
\textbf{Fragment} &\textbf{Length} & \textbf{Width} & \textbf{Thickness} \\
\hline
Pumpkin   & $12.4\pm 0.8$ & $6.8\pm 0.3$ & $2.4\pm 0.3$\\
Raisins   & $12.3\pm 1.5$ & $5.9\pm 1.1$ & $4.2\pm 0.5$\\
\end{tabular}
\caption{\small 
{\sc Fragment Dimensions.} In millimeters, the average length (longer axis), width (shorter axis), and height of ten instances of each type, $\pm$ one standard deviation.
}
\small
\label{tab:dimensions}
\end{table}

\noindent \textbf{Setup of dVRK.} The setup is shown in Figure~\ref{fig:dvrk_setup}. We use one dVRK arm with a gripper end-effector, called a \emph{large needle driver} (see Figure~\ref{fig:yaw_pitch_roll}), that can be opened up to 75\degree, or a gripper width of 10mm. We assume access only to the perception available in current da Vinci deployments: a 1920x1080 stereo endoscopic camera. These are located above a $7.5\mathrm{cm}\times 7.5\mathrm{cm}$ flat elastic silicone-based tissue phantom. We do not assume access to a motion capture system as in~\cite{mahler2014case} or force sensing as in~\cite{okamura2004}. All grasps during debridement are pinch grasps. Finally, we assume that at any time we can query the current $x_b$ and $\phi$ of the end-effector, and the two endoscopic lenses to obtain pixels of a world point, from which we can infer $x_c$ using disparity.

\noindent \textbf{Debridement.} We assume that there are initially eight fragments on a white sheet of paper on the tissue phantom, and that all are small enough to be gripped by a fully open gripper in the correct position and orientation. While the fragments may be of arbitrary orientation, we assume that they are non-overlapping with a minimum of 3mm of space between any two fragments. As fragments, we use pumpkin seeds and raisins with sizes shown in Table~\ref{tab:dimensions}.

\begin{figure*}[t]
\centering
\includegraphics[width=0.90\textwidth]{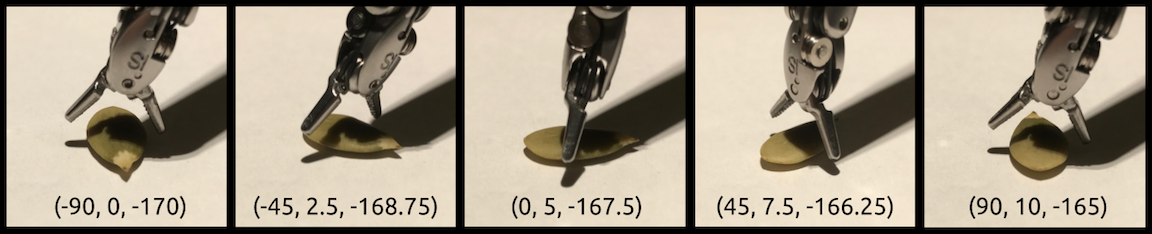}
\caption{\small
For each fragment, we compute its angle of orientation (by fitting an ellipse) and approximate it with the nearest yaw value in our discretization: -90\degree, -45\degree, 0\degree, 45\degree, 90\degree. Each yields hand-tuned pitch and roll values that we observed empirically allowed both tips to touch the workspace simultaneously when lowering the end-effector to the surface. Above, the pumpkin seed is rotated counter-clockwise across consecutive images, and the resulting yaw, along with our chosen pitch and roll values, are listed below (see Figure~\ref{fig:yaw_pitch_roll}). For all five, moving downwards (adjusting $b_z$ only) a few millimeters is sufficient to get the gripper in a spot where it can reliably grasp the seed.
}
\label{fig:rotations_interpolated}
\end{figure*}

\noindent \textbf{Rotations.} To maintain reliability of debridement, we constrain the dVRK gripper tips to touch the workstation plane with fixed ranges of rotation. The yaw values need the full range $[-90\degree,90\degree]$ so that the gripper can reliably grasp arbitrarily-oriented fragments in their minor elliptical axis. Thus, we make the simplifying assumption that the pitch and roll can be known \emph{conditioned} on a yaw.

We additionally assume that before debridement, we can use OpenCV to fit an ellipse to all fragments to get an angle of orientation along the major axis. This angle can be translated into a yaw value for the dVRK.  We furthermore discretize the yaw values into five choices, -90\degree, -45\degree, 0\degree, 45\degree, and 90\degree. Each value corresponds to a distinct training data for non-linear regressors, which we describe in Section~\ref{ssec:fine}.\footnote{While -90\degree and 90\degree technically represent the same world space angle, we observed that the dVRK needs slightly different pitch and roll values (see Figure~\ref{fig:rotations_interpolated}) for these cases to comfortably grip the fragments.} During debridement applications, each elliptical angle is mapped to the nearest yaw value, and then pitch and roll are determined from that (see Figure~\ref{fig:rotations_interpolated}).

\subsection{Evaluation}\label{ssec:evaluate_calibration}

\begin{figure}[t]
\centering
\includegraphics[width=0.40\textwidth]{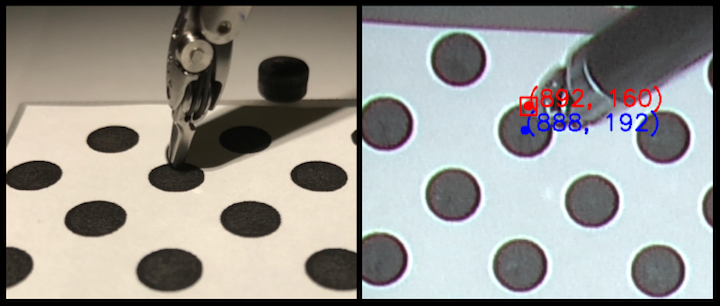}
\caption{\small
Evaluating calibration methods. Left: the end-effector after the dVRK attempts to touch the center of a target circle. Right: calibration error is the pixel-wise distance between the tip of the closed end-effector (colored red, pixel location (892,160)) and the center of the target circle (colored blue, pixel location (888,192)).
}
\label{fig:evaluate_calibration}
\end{figure}

\noindent \textbf{Calibration.} To compare calibration algorithms, we use a printed grid with black circles overlaid on our silicone tissue. For each circle, we get the left and right camera pixels of its center via OpenCV contour detection, and with the disparity, we can determine $x_c$. The calibration provides us with $x_b$, and the dVRK's closed end-effector moves to $x_b$. A human then identifies the location of the end-effector in pixel space. The distance in pixels between the circle center and the end-effector tip (i.e., its lowest point) is the calibration error, and the goal is to minimize this quantity. See Figure~\ref{fig:evaluate_calibration} for details.

\noindent \textbf{Debridement.} Our objective is to debride as many fragments as possible while minimizing the time per fragment. A fragment is \emph{successfully debrided} when the dVRK is able to grasp it and move it above a receptacle outside of the workspace without collisions with other fragments, while using both grippers. We defer a detailed discussion on the debridement setup to Section~\ref{ssec:debridement_setup_and_fails}.

\section{Methodology}\label{sec:method}

Our method has two phases. The first phase quickly and automatically obtains a large amount of coarse data. The second phase uses the results of Phase I and human intervention to generate a small amount of high quality data.

\subsection{Phase I: Coarse Motion Bias with DNN}\label{ssec:coarse}

The goal in this phase is to collect the dataset
\begin{equation}\label{eq:data_dnn}
\mathcal{X}_{\rm DNN} = \{ ((x_c,\phi)^{(1)}, x_b^{(1)}), \ldots, ((x_c,\phi)^{(N)},x_b^{(N)}) \},
\end{equation}
where each $(x_c,\phi)^{(i)}$ is the concatenation of a camera position and a rotation vector and $x_b^{(i)}$ is a base position. The rotation is part of the \emph{input}, which follows from our rotation assumptions in Section~\ref{sec:problem_statement} since the $\phi_y$ values (and consequently, $\phi_p$ and $\phi_r$) are determined using OpenCV.

To build $\mathcal{X}_{\rm DNN}$, we automatically execute trajectories of the dVRK. For each trajectory, its starting location is one of the four workspace corners. Then, its target position $(b_x,b_y,b_z)$ is randomly chosen, though each coordinate is constrained within pre-determined safe ranges (e.g. $b_z$ is restricted to be level with the workspace). 

Once the starting and target locations are set, each trajectory is split into a series of shorter, linear movements, involving an adjustment in the end-effector position of about 1mm. After each of these shorter movements, we pause the trajectory and record the position $x_b$ and rotation $\phi$.

Forming $\mathcal{X}_{\rm DNN}$ also requires the camera position $x_c$ at these points, so before executing the trajectories we apply red tape to the end-effector, thus allowing the dVRK to use HSV thresholding to detect the location of red contours. This is a substantially cheaper alternative than the sensors used in~\cite{mahler2014case}, though it is less reliable, because (as seen in Figure~\ref{fig:dvrk_setup}), the cameras are oriented directly above the workspace, and the dVRK's wrist can block the red tape from view. As we show in Section~\ref{ssec:dnn}, we need to perform data cleaning to exclude these cases from $\mathcal{X}_{\rm DNN}$.

Consequently, to increase the set of candidate data points for $\mathcal{X}_{\rm DNN}$ prior to cleaning, at each time we paused the trajectory, we additionally executed three random rotations at the same robot position, with $\phi_y,\phi_p$ and $\phi_r$ values randomly chosen in $[-90\degree,90\degree]$, $[-15\degree,25\degree]$ and $[-180\degree,-150\degree]$.

Once the data $\mathcal{X}_{\rm DNN}$ is collected, we can use it to train a Deep Neural Network $f_{\rm DNN} : \mathbb{R}^6 \to \mathbb{R}^3$ to map $(c_x, c_y, c_z, \phi_y, \phi_p, \phi_r)$ to a predicted $(b_x,b_y,b_z)$.

\subsection{Phase II: Fine Motion Bias with RF}\label{ssec:fine}

\begin{figure}[t]
\centering
\includegraphics[width=0.40\textwidth]{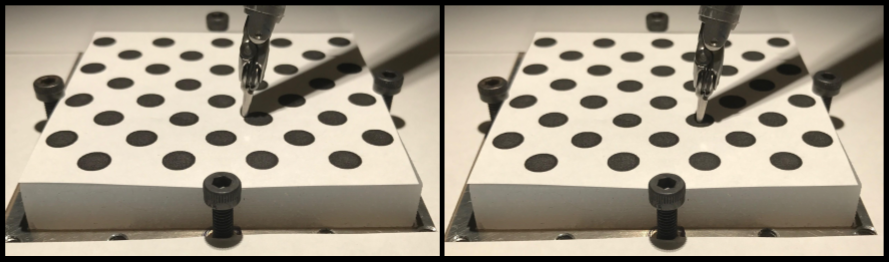}
\caption{\small
Fine-tuning the calibration. Left: the dVRK's closed end-effector attempts to move to the center of a circle via $f_{\rm DNN}$, but it is off by several millimeters. Right: a human moves the end-effector through direct contact to the center of the circle, and the world-space transformation is stored as training data.
}
\label{fig:human_guided_calibration}
\end{figure}

In our second phase, we collect five datasets:
\begin{equation}
\mathcal{X}_{\rm RF}^{(\phi_y)} = \{ (x_b^{(1)},\epsilon^{(1)}), \ldots, (x_b^{(N)},\epsilon^{(N)}) \},
\end{equation}
for each $\phi_y \in \{-90\degree,-45\degree,0\degree,45\degree,90\degree\}$, where each $x_b^{(i)}$ and $\epsilon^{(i)}$ refer to a predicted base position and a residual error vector, respectively.

The rationale for this second phase is that, after obtaining $f_{\rm DNN}$, the calibration may still be imperfect due to noisy data (e.g. slight perception mismatches when locating the red tape among the two cameras). Since our debridement scenarios require fine-grained calibration, as in~\cite{pastor2013,mahler2014case}, we model the residuals using a non-linear function.

Figure~\ref{fig:human_guided_calibration} visualizes a step in this process. We use the same printed calibration grid of black circles, with the centers known via OpenCV contour detection. From our Section~\ref{sec:problem_statement} assumptions, we have five discretized yaws and thus five rotation vectors $\phi$ (see Figure~\ref{fig:rotations_interpolated}). For each $\phi$, we command the dVRK's closed end-effector to touch the center of each circle in the grid while maintaining that rotation. All rotations use the same calibration function $f_{\rm DNN}$ to get $x_b$ from $(x_c,\phi)$.

In this process, the dVRK's end-effector tip may not reach the exact center of the circles, as judged by a human. Consequently, for each circle, once the dVRK's end-effector has stopped moving, a human directly (not through teleoperation) adjusts it to the center of the circle. For the $i$th circle of yaw $\phi_y$, we record the world-space transformation and store it in the data $\mathcal{X}_{\rm RF}^{(\phi_y)}$ as $\epsilon^{(i)}$.

With these datasets, it is then possible to train five random forests $f_{\rm RF}^{(\phi_y)} : \mathbb{R}^3 \to \mathbb{R}^3$, each of which maps \emph{a predicted} $(b_x, b_y, b_z)$ from $f_{\rm DNN}$ to a residual vector $(\epsilon_x,\epsilon_y,\epsilon_z)$. 

For a given fragment in the workspace, we can detect $(c_x,c_y,c_z)$ and its orientation angle. The orientation angle then maps to the nearest yaw value in $\{-90\degree,-45\degree,0\degree,45\degree,90\degree\}$, and thus the pre-computed pitch and roll values. The result from calibration is the mapping from $(x_c,\phi) = (c_x,c_y,c_z,\phi_y,\phi_p,\phi_r)$ to $x_b$ as:
\begin{equation}\label{eq:two_step_pipeline}
f_{\rm DNN}(x_c, \phi) + f_{\rm RF}^{(\phi_y)}(f_{\rm DNN}(x_c, \phi)) = x_b.
\end{equation}

%
%
\begin{figure*}[t]
\centering
\includegraphics[width=0.90\textwidth]{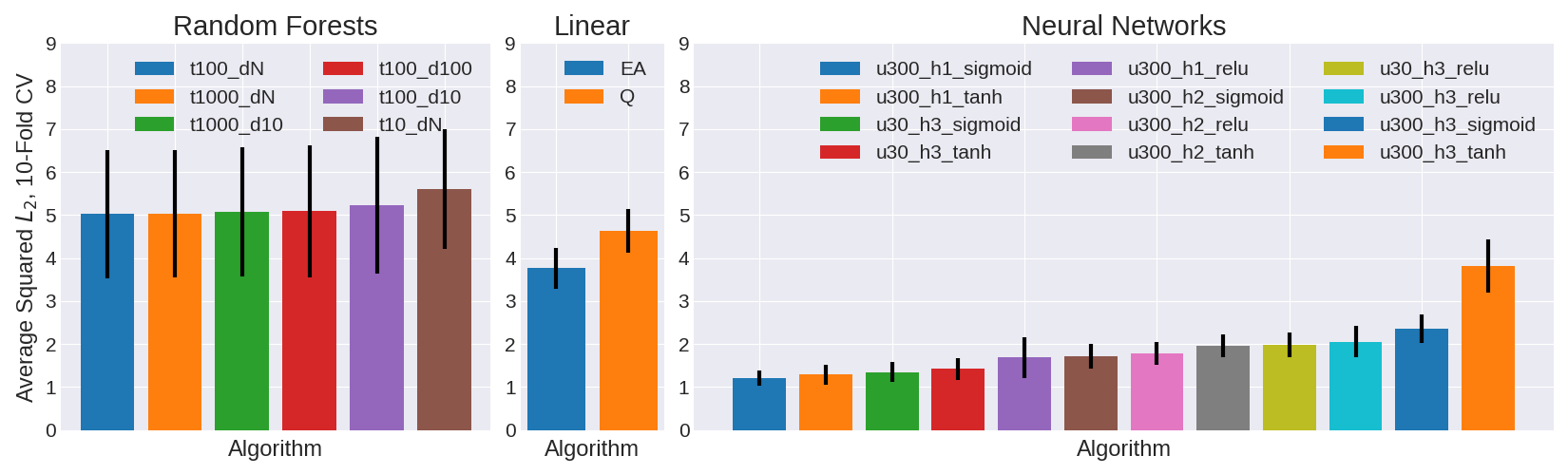}
\caption{\small
Average squared $L_2$ loss of different algorithms on the data from Phase I (see Section~\ref{ssec:coarse}), each taken over a 10-fold cross validation with one standard deviation shown with an error bar. Left: Random Forest regressors. Key: \texttt{tA\_dB} where \texttt{A} is the number of trees and \texttt{B} is the depth (or none in the case of \texttt{N}, which is the default setting of \texttt{scikit-learn}~\cite{scikit-learn}). Middle: linear regression. Key: \texttt{EA} and \texttt{Q} means angles are Euler angles or quaternions, respectively. Right: neural networks (top 12 out of 24). Key: \texttt{uA\_hB\_C} where \texttt{A} is the number of units in each hidden layer, \texttt{B} is the number of hidden layers, and \texttt{C} is the activation.
}
\label{fig:testing_different_algos}
\end{figure*}

\section{Calibration Experiments}\label{sec:calibration_experiments}

\subsection{Phase I}\label{ssec:dnn}

We ran the dVRK arm for 57 trajectories, which collectively provided 5256 initial data points $\{((x_c,\phi), x_b)^{(i)}\}_{i=1}^{5256}$. Different trajectories have different lengths, and they contribute to this dataset in proportion to their length --- the longer the trajectory, the more data. We cleaned the data by keeping those points where the left and right cameras both detected a pixel center (of the red tape) corresponding to the same workspace location, resulting in 1939 examples. 

To determine the architecture for $f_{\rm DNN}$, we ran a preliminary search over several hyperparameter settings for fully connected networks, and settled on three hidden layers with 300 units each, and with the ReLU activation~\cite{Nair2010}. After running experiments with this network, we later ran more extensive hyperparameter searches by varying the number of layers (1-4), number of hidden units in each layer (30 each or 300 each) and the nonlinearity (ReLU, sigmoid, or tanh). All training used Adam~\cite{adam2014} on the squared $L_2$ loss.

Figure~\ref{fig:testing_different_algos} shows the performance of the top 12 of these 24 hyperparameter settings, reported as average squared $L_2$ losses over a 10-fold cross validation of $\mathcal{X}_{\rm DNN}$ after 1000 training epochs. We additionally benchmarked against linear regression (including the Euler angle vs quaternion formulation) and Random Forest regression. The architecture in our experiments obtains an average squared $L_2$ loss of $2.06\mathrm{mm}^2$. Several other networks, however, have lower loss values, with the best one using one hidden layer of 300 units with the sigmoid. This obtains a loss of $1.21\mathrm{mm}^2$. We will explore this and similar alternatives in future work.


\subsection{Calibration Performance}\label{ssec:calibration}

\begin{table}[t]
\centering
\small
\begin{tabular}{l r r r r}
\textbf{Mapping} & \textbf{Yaw} $\phi_y$ &\textbf{Mean $\pm$ SE} & \textbf{Med.} & \textbf{(Min, Max)} \\
\hline
RBT & -90  & $39.9\pm 1.8$ & 36.8 & (26.4, 60.0)  \\
RBT & -45  & $53.6\pm 1.5$ & 50.7 & (43.2, 78.1)  \\
RBT & 0    & $52.0\pm 2.6$ & 53.7 & (29.0, 85.7) \\
RBT & 45   & $72.0\pm 5.3$ & 66.4 & (32.0, 167.2) \\
RBT & 90   & $44.1\pm 1.4$ & 43.1 & (29.1, 67.2) \\
RBT & [-90,90] & $47.1\pm 2.1$ & 48.3 & (19.7, 75.1)  \\
\hline
DNN & -90  & $26.7\pm 1.5$ & 22.0 & (12.7, 47.8)  \\
DNN & -45  & $28.3\pm 0.8$ & 29.4 & (18.3, 38.9)  \\
DNN &   0  & $22.7\pm 1.0$ & 22.6 & (11.4, 39.4) \\
DNN &  45  & $21.5\pm 1.3$ & 21.0 & (7.0, 39.1) \\
DNN &  90  & $22.5\pm 2.3$ & 25.0 & (2.2, 53.3) \\
DNN & [-90,90] & $23.1\pm 1.6$ & 21.0 & (3.6, 48.8)  \\
\hline
DNN+RF & -90 & $10.5\pm 1.0$ &  8.0 & (1.0, 25.0) \\
DNN+RF & -45 & $14.2\pm 1.1$ & 14.1 & (1.0, 29.1) \\
DNN+RF &   0 & $12.8\pm 1.2$ & 12.0 & (1.4, 27.2) \\
DNN+RF &  45 & $12.1\pm 1.2$ & 11.6 & (2.0, 30.0) \\
DNN+RF &  90 & $ 8.9\pm 1.0$ &  8.4 & (1.0, 28.0) \\
DNN+RF & [-90,90] & $14.7\pm 1.4$ & 13.0 & (2.2, 34.9) \\
\hline
\end{tabular}
\caption{\small 
{\sc Calibration Error Results (In Pixels).} The first six are RBT estimators. The next six are based on a DNN, and the last six combine the DNN with RFs. These are conditioned on yaw values. For each mapping, we evaluated its accuracy by measuring the pixel-wise error distance from the dVRK's closed end-effector to the center of the circles in our calibration grid (see Section~\ref{sec:problem_statement}). Error values are in pixels and based on 35 distances, one for each circle in the printed calibration grid. As stated in Section~\ref{ssec:calibration}, 1mm in the workspace is roughly 11.3 pixels. 
}
\label{tab:calibration_benchmarks}
\end{table}

We benchmark the calibration performance of various methods by following the procedure in Section~\ref{ssec:evaluate_calibration}: commanding the dVRK end-effector to reach the center of each circle in the calibration grid and measuring how far the result is from the true center. We test the following:

\begin{itemize}
\item \textbf{Rigid Body Transformation (RBT)}. Following~\cite{mahler2014case}, we initially test a rigid body transformation where we map the camera to base positions by means of a rotation and a translation.\footnote{For the RBT, we divide the data of 1939 elements into five groups, one for each of the yaw values. Each data point is assigned to the group with the closest yaw to its actual $\phi_y$.} We determine the rigid body by solving an orthogonally constrained least squares problem with singular value decomposition~\cite{Hartley2003}.
\item \textbf{Deep Neural Network (DNN)}. Using only the function $f_{\rm DNN}$ (see Section~\ref{ssec:coarse}).
\item \textbf{DNN and Random Forest (DNN+RF)}. Using a DNN and then a Random Forest, i.e. Equation~\ref{eq:two_step_pipeline}. 
We used 100 trees for each, with no specified depth, the default behavior of \texttt{scikit-learn}~\cite{scikit-learn}.
\end{itemize}

We test using all yaws in our discretization. In addition, we add a sixth ``yaw setting'' where, for each circle, we randomly choose a yaw value in $[-90,90]$, so that the dVRK maps each random yaw into the closest value in the discretization.

Table~\ref{tab:calibration_benchmarks} lists the errors (in \emph{pixels}) based on all 35 circle centers of the above algorithms. The RBT exhibits substantial calibration errors, which is likely due to coarse data. The average pixel error among the six settings is 51.45. The DNN performs substantially better with improvements across the entire set of $\phi_y$ choices, roughly halving the error in pixel space (24.13 pixels). Adding Random Forests halves the errors again (12.20 pixels). The standard error of the mean implies that the differences are statistically significant.

We measured 1mm in various locations of the workspace in terms of pixels and found the average correspondence to be roughly 1mm to 11.3 pixels. This is distance in $(b_x,b_y)$ only, not including $b_z$ since the vertical distance can't be captured using a single camera image.\footnote{Errors in $b_z$ are typically less problematic than errors in $b_x$ and $b_y$ because the tissue phantom is elastic and the gripper can press a few millimeters in it. We also fine-tune a vertical offset beforehand.} The RBT, DNN, and DNN+RF workspace errors are roughly 4.55mm, 2.14mm, and 1.08mm, respectively.

\section{Debridement Experiments}\label{sec:debridement_experiments}


\subsection{Debridement Setup and Failure Modes}\label{ssec:debridement_setup_and_fails}

\begin{figure*}[t]
\centering
\includegraphics[width=0.90\textwidth]{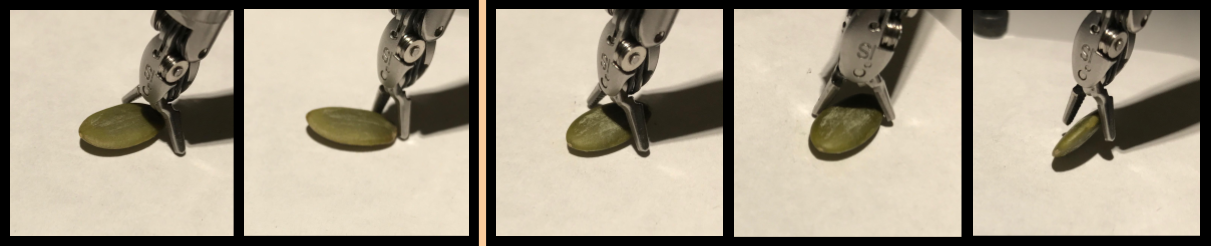}
\caption{\small
Properties of pumpkin seeds during debridement. These seeds slide and rotate on the paper surface. Type A errors frequently result in the seed getting pushed out of the gripper's control (first two images). This can also happen even if calibration is accurate and the dVRK ends up in a reasonable location. The third image shows a gripper initially in an excellent spot to grab the seed, but one tip can rotate the seed before it gets gripped (fourth image) which can cause it to snap out of control (fifth image). This is a type B error.
}
\label{fig:pumpkin_seeds}
\end{figure*}

\begin{figure}[t]
\centering
\includegraphics[width=0.45\textwidth]{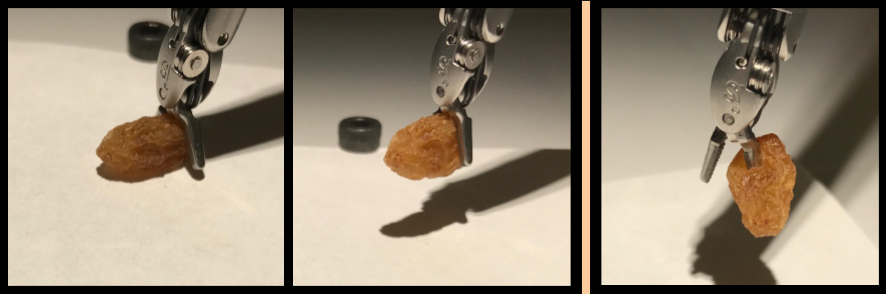}
\caption{\small
Properties of raisins during debridement. Raisins are sticky and elastic, and a gripper that grips it by the edge often pulls it successfully out of the workspace (first two images). A type C error is when the gripper pulls it out using one tip (third image).
}
\label{fig:raisins}
\end{figure}

We benchmark DNN and DNN+RF on debridement tasks using pumpkin seeds and raisins. (We do not test the RBT since its calibration performance is substantially worse than the DNN alone.) We follow the setup described in Section~\ref{sec:problem_statement}. Prior to each trial, we inspected the contours of all eight fragments to check that perception was accurate (i.e., that their centers were correctly identified). For each fragment, a human judge classified its debridement outcome into one of four possibilities:

\begin{itemize}
\item \textbf{Success}: When a fragment is pulled out of the workspace using both grippers.

\item \textbf{Error Type ``A''}: True calibration errors. The gripper ends up too far from the center of a fragment to successfully grasp it. These are the most critical sources of error, as they suggest flaws in the calibration procedure.

\item \textbf{Error Type ``B''}: A fragment was \emph{not} pulled out of the workspace despite accurate calibration. These happen with pumpkin seeds, which are smooth and lack friction with the paper surface. They can therefore slide out of control of the gripper (see Figure~\ref{fig:pumpkin_seeds}).

\item \textbf{Error Type ``C''}: When a gripper ``grabs'' a raisin with just one tip, implying that the fragment is pulled out of the workspace despite \emph{inaccurate} calibration. See Figure~\ref{fig:raisins} for details. We do not consider the case when the dVRK grabs a raisin off-center using \emph{both} tips, as we consider this a more realistic debridement since it does not involve puncturing a fragment. 
\end{itemize}

The rationale for splitting outcomes into these cases is that the ``Success'' metric provides a clear number as to how many fragments were pulled out of the workspace without getting punctured (i.e., excluding type C). In addition, distinguishing between type A and B errors makes it clearer to identify if failures are due to calibration or properties of the fragment, gripper, and/or setup.

For each trial, the same human judge labeled the result of all eight fragments. Certain failure cases were ambiguous (e.g., may have been A or B), so to be conservative, \emph{all ambiguous cases involving type A were labeled as type A}.

One limitation of this setup is that evaluating the debridement outcome is subjective. To partially mitigate this, we ran our trials in a \emph{blind} fashion: the choice of whether DNN or DNN+RF was used for calibration was \emph{unknown} to the human judge until after labeling the outcome (success, A, B, or C) of all eight seeds in a trial.

\subsection{Debridement Results}\label{ssec:debridement_results}

%
\begin{table}[t]
\small
\centering
\begin{tabular}{l l l}
\textbf{Trial} & \textbf{DNN} & \textbf{DNN+RF} \\
\hline
1  & \texttt{-,A,A,A,B,-,-,-}  & \texttt{-,-,-,-,-,-,-,B} \\
2  & \texttt{-,-,-,A,-,-,-,-}  & \texttt{-,B,-,-,-,-,-,B} \\
3  & \texttt{-,A,A,A,A,-,-,-}  & \texttt{-,-,-,-,-,-,-,-} \\
4  & \texttt{A,-,B,-,-,-,-,-}  & \texttt{-,-,-,-,-,-,-,B} \\
5  & \texttt{-,A,B,A,-,-,A,-}  & \texttt{-,-,-,-,-,-,-,-} \\
6  & \texttt{A,B,-,A,-,B,-,-}  & \texttt{-,-,B,-,-,-,-,-} \\
7  & \texttt{-,A,A,A,-,-,B,-}  & \texttt{-,-,-,-,-,-,-,-} \\
8  & \texttt{A,-,-,-,-,-,B,-}  & \texttt{B,-,-,-,-,-,-,-} \\
9  & \texttt{A,-,A,A,-,-,-,-}  & \texttt{-,-,-,-,-,-,-,-} \\
10 & \texttt{-,B,-,-,-,-,B,-}  & \texttt{-,-,-,-,-,-,-,-} \\
11 & \texttt{-,-,-,A,A,A,-,-}  & \texttt{-,-,-,-,-,-,-,-} \\
12 & \texttt{-,A,A,-,-,B,-,-}  & \texttt{-,-,-,B,-,-,-,-} \\
13 & \texttt{A,-,-,A,-,-,A,-}  & \texttt{-,B,-,-,-,-,-,-} \\
14 & \texttt{A,B,A,A,A,A,-,-}  & \texttt{-,-,-,-,-,-,A,-} \\
15 & \texttt{-,B,-,-,-,-,B,-}  & \texttt{-,B,-,-,-,-,-,-} \\
\hline
Count   & \texttt{A:34, B:13, C:0} & \texttt{A:1, B:9, C:0} \\
Success & 73/120                   & 110/120 \\
\hline
\end{tabular}
\caption{\small
{\sc Debridement Using Pumpkin Seeds.} Results of 15 trials with eight pumpkin seeds each, for the DNN and DNN+RF cases following the setup in Section~\ref{ssec:debridement_setup_and_fails}. The columns indicate the outcome of \emph{each} fragment, separated by commas. A dash (\texttt{-}) indicates a fragment that was successfully pulled out of the workspace and the ``Success'' row indicates the total across 15 trials for each setting. Any other occurrence (\texttt{A}, \texttt{B}, or \texttt{C}) follows the labeling as described in Section~\ref{ssec:debridement_setup_and_fails}. The average running time for these 30 trials was $57.62\pm 1.17$ seconds.
}
\label{tab:pumpkin_seeds}
\end{table}

We ran 15 trials for each fragment. Tables~\ref{tab:pumpkin_seeds} and~\ref{tab:raisins} list the debridement results for pumpkin seeds and raisins, respectively. In particular, they contain the exact outcome of each fragment and a tally of the error cases.

The results suggest that our DNN+RF calibration is highly reliable. Overall success rates are 110/120 (91.67\%) and 119/120 (99.17\%) for pumpkin seeds and raisins, respectively. These exceed the 89.4\% success rate in the single-arm scenario reported in~\cite{Kehoe2014}; \cite{mahler2014case} did not report exact success rates but said results were similar to those in~\cite{Kehoe2014}.

Failure cases due to calibration errors (type A) occur just once out of 120 instances for both pumpkin seeds and raisins (0.83\% of the time). These are lower than the type A error rates with the DNN only, which are 34/120 (28.33\%) for pumpkin seeds and 4/120 (3.33\%) for raisins. The difference in type A error rates in the DNN case for the pumpkin seeds versus raisins is due to how raisins are elastic and can be pulled successfully despite slightly inaccurate calibration.

%
\begin{table}[t]
\small
\centering
\begin{tabular}{l l l}
\textbf{Trial} & \textbf{DNN} & \textbf{DNN+RF} \\
\hline
1  & \texttt{-,-,-,-,-,-,-,-}  & \texttt{-,-,-,-,-,-,-,-} \\
2  & \texttt{-,-,-,-,-,-,-,-}  & \texttt{-,-,-,-,-,-,-,-} \\
3  & \texttt{-,-,-,-,-,-,-,-}  & \texttt{-,-,-,-,-,-,-,-} \\
4  & \texttt{-,-,-,-,-,A,-,-}  & \texttt{-,-,-,-,-,-,-,-} \\
5  & \texttt{-,-,-,-,C,-,-,-}  & \texttt{-,-,-,-,-,-,-,-} \\
6  & \texttt{-,-,-,-,-,-,-,-}  & \texttt{-,-,-,-,-,-,-,-} \\
7  & \texttt{-,-,-,-,C,-,-,-}  & \texttt{-,-,-,-,-,-,-,-} \\
8  & \texttt{-,-,-,-,-,-,-,-}  & \texttt{-,-,-,-,-,-,-,-} \\
9  & \texttt{-,-,A,-,-,-,-,-}  & \texttt{-,-,-,-,-,-,-,-} \\
10 & \texttt{-,-,-,-,-,-,-,-}  & \texttt{-,-,-,-,-,-,-,-} \\
11 & \texttt{-,-,-,A,-,-,-,-}  & \texttt{-,-,-,-,-,-,-,-} \\
12 & \texttt{-,-,-,-,-,-,-,-}  & \texttt{-,-,-,-,-,-,-,-} \\
13 & \texttt{A,-,-,-,-,-,-,-}  & \texttt{-,-,-,-,-,-,-,-} \\
14 & \texttt{-,-,-,C,-,-,-,-}  & \texttt{-,-,-,-,-,-,A,-} \\
15 & \texttt{-,-,-,-,-,-,-,-}  & \texttt{-,-,-,-,-,-,-,-} \\
\hline
Count   & \texttt{A:4, B:0, C:3} & \texttt{A:1, B:0, C:0} \\
Success & 113/120                & 119/120 \\
\hline
\end{tabular}
\caption{\small
{\sc Debridement Using Raisins.} Results following the same setup and description as in Table~\ref{tab:pumpkin_seeds}, except that these use raisins instead of pumpkin seeds as fragments. The average running time for these 30 trials was $58.57\pm 0.80$ seconds.
}
\label{tab:raisins}
\end{table}

We observe that type B errors only occur with pumpkin seeds: 13 cases for DNN and 9 cases for DNN+RF, but zero times for both raisin scenarios. Again, this follows from properties of the fragments, workspace, and gripper. In particular, most of the type B errors occurred when the seeds had not been part of previous trials and thus retained their natural smoothness. When the gripper grabs seeds, it slightly deforms their edges, making them coarser and less likely to slip out of control in future trials. To avoid this from excessively biasing our results, we replaced seeds every 5 trials or after they experienced any significant damage.

Error type C, for raisins, occurred three times for DNN but zero times for DNN+RF, providing some additional evidence that the extra RF correction is beneficial.

Average runtimes for each trial were 57.62 seconds for pumpkin seeds and 58.57 seconds for raisins. These correspond to debridement rates of 7.20 seconds for pumpkin seeds and 7.32 seconds for raisins, more than an order of magnitude faster than the 91.8 seconds reported in~\cite{Kehoe2014} and more than 2.1x faster compared to the 15.8 seconds from~\cite{mahler2014case} who furthermore tested only the faster two-arm scenario. 

To investigate runtime in more detail, we ran a frame-by-frame analysis of a 25fps video of a pumpkin seed trial. We considered six phases for each fragment: \emph{to seed}, \emph{lower}, \emph{grip}, \emph{raise}, \emph{to receptacle}, and \emph{drop \& rotate}. The average frame count for each phase across the eight seeds was, respectively, 26.0, 16.1, 37.5, 24.0, 42.1, and 38.5. Moving back to the receptacle (\emph{to receptacle}), opening the gripper to drop the seed and then rotating (\emph{drop \& rotate}) for the next seed are the slowest phases. We had to consider these as distinct steps because our dVRK arm moves faster if it maintains a fixed $\phi$ while moving. A possible speed-up could involve figuring out how to merge the two steps, rotating while moving back to the receptacle, without loss in speed.

\section{Conclusions}


This paper proposes and evaluates a novel two-phase coarse-to-fine calibration method that combines a DNN with a Random Forest to learn compensation bias. Experiments suggest that this method can increase success rates for debridement over prior results with much larger fragments and achieving a 2x speedup.  In future work we will apply the method to other tasks and explore more efficient data collection methods and avoiding the need to hand-tune and discretize rotations. Other possibilities for future work include future study of the DNN and RF combination and alternatives, using more elaborate debridement setups with clutter, and experiments with subtasks such as suturing.

{\scriptsize
\section*{Acknowledgments}
This research was performed at the AUTOLAB at UC Berkeley in affiliation with the RISE Lab, BAIR, and the CITRIS "People and Robots" (CPAR) Initiative: http://robotics.citris-uc.org in affiliation with UC Berkeley's Center for Automation and Learning for Medical Robotics (Cal-MR). The authors were supported in part by the U.S. National Science Foundation under NRI Award IIS-1227536: Multilateral Manipulation by Human-Robot Collaborative Systems, and by Google, UC Berkeley's Algorithms, Machines, and People Lab, and by a major equipment grant from Intuitive Surgical. Daniel Seita is supported by a National Physical Science Consortium Fellowship. We thank Nate Armstrong, Michael Laskey, and Jeffrey Mahler.}


\begingroup
\renewbibmacro{pageref}{}
\printbibliography
\endgroup

\end{document}